\documentclass[11pt,a4paper]{article}
\usepackage[hyperref]{acl2021}
\usepackage{times}
\usepackage{latexsym}

\usepackage[algo2e,ruled,vlined]{algorithm2e}
\usepackage{graphicx}
\usepackage[bottom]{footmisc}
\usepackage{booktabs}
\usepackage{soul}
\usepackage{algorithm}
\usepackage{algorithm2e}
\usepackage{algorithmic}
\usepackage{multirow, multicol}
\usepackage{amsmath}
\usepackage{amsfonts}
\usepackage{placeins}
\usepackage{tabularx}
\usepackage{adjustbox}
\usepackage{xcolor}
\usepackage{array, makecell} 
\usepackage{arydshln}
\usepackage{float}

\def\[#1\]{\begin{align}#1\end{align}}

\usepackage{microtype}

\aclfinalcopy 
\def\aclpaperid{1972} 

\newcommand\blfootnote[1]{%
\begingroup
\renewcommand\thefootnote{}\footnote{#1}%
\addtocounter{footnote}{-1}%
\endgroup
}

\title{Improving Lexically Constrained Neural Machine Translation \\ with Source-Conditioned Masked Span Prediction}

\author{Gyubok Lee\textsuperscript{*} \qquad Seongjun Yang\textsuperscript{*} \qquad Edward Choi\\
Graduate School of AI, KAIST \\
\texttt{\{gyubok.lee,seongjunyang,edwardchoi\}@kaist.ac.kr} \\}

\begin{document}
\maketitle
\begin{abstract}
\blfootnote{* equal contributions}

Accurate terminology translation is crucial for ensuring the practicality and reliability of neural machine translation (NMT) systems. 
To address this, lexically constrained NMT explores various methods to ensure pre-specified words and phrases appear in the translation output.
However, in many cases, those methods are studied on general domain corpora, where the terms are mostly uni- and bi-grams ($>$98\%).
In this paper, we instead tackle a more challenging setup consisting of domain-specific corpora with much longer n-gram and highly specialized terms. Inspired by the recent success of masked span prediction models, we propose a simple and effective training strategy that achieves consistent improvements on both terminology and sentence-level translation for three domain-specific corpora in two language pairs.


\end{abstract}

\section{Introduction}
\label{sec:introduction}
Despite its recent success in neural machine translation (NMT) \cite{wu2016google, johnson2017google, barrault2020findings}, delivering correct terms in the translation output is still a vital component for high-quality translation. This concern becomes more salient in domain-specific scenarios, such as in legal documents, where generating correct and consistent terminology is key to ensuring the practicality and reliability of machine translation (MT) systems \cite{chu2018survey, exel2020terminology}.

\begin{figure}[t!]
\centering
\includegraphics[width=7.8cm]{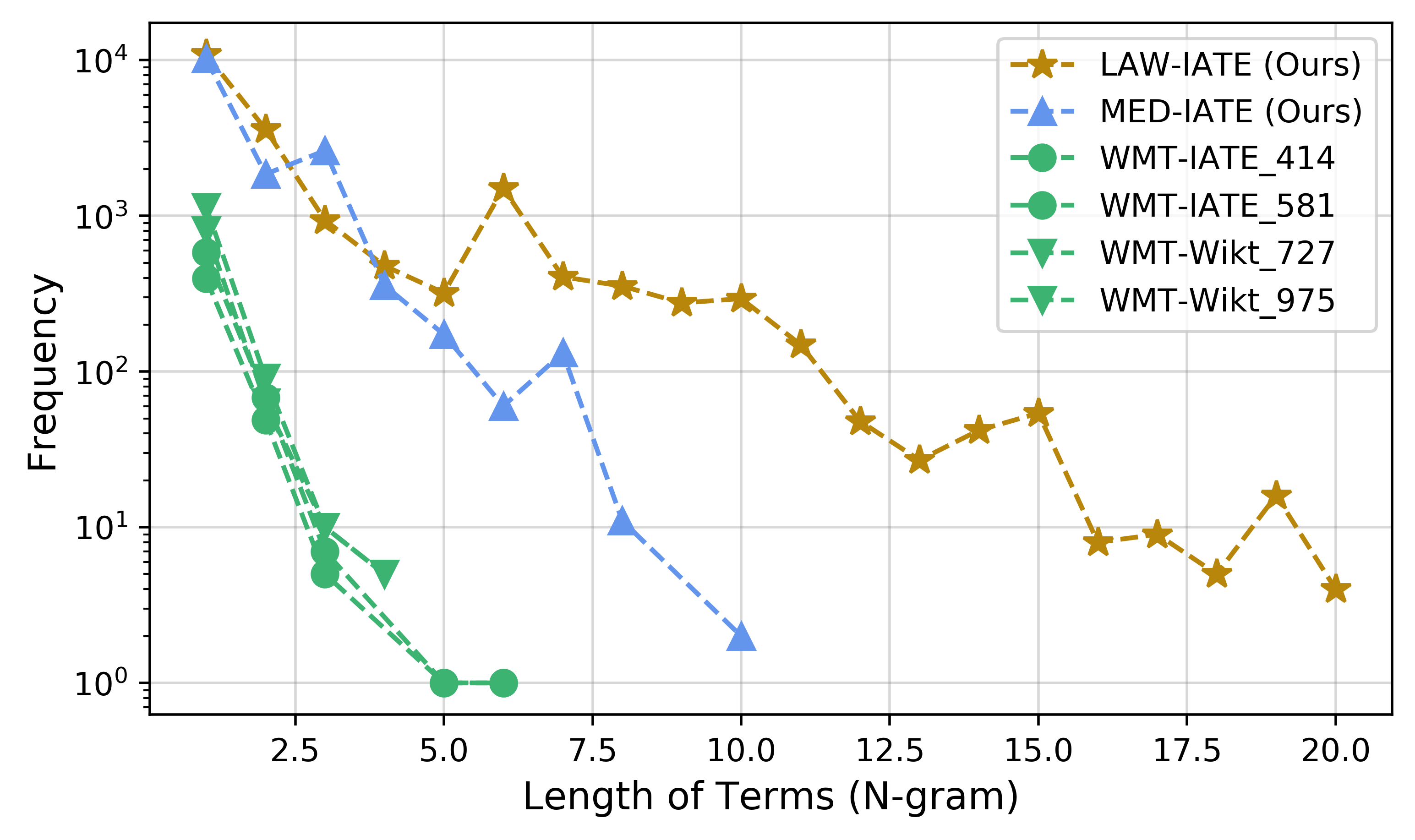}
 \caption{The frequency of terms sorted by n-gram between \citet{dinu2019training}'s and our test splits. While the terms in WMT De-En are mostly uni- or bi-grams, our setup contains heavy-tailed n-gram distributions with more quantity and diversity in terminology.}
\label{fig:ngram_dist}
\end{figure}

To address this, lexically constrained NMT works have proposed various methods to preserve terminology in translations as lexical constraints with or without the help of a term dictionary at test time. In most lexically constrained NMT setups, datasets and terms used for training and evaluating the methods are extracted from WMT news corpora \cite{dinu2019training, susanto2020lexically, chen2020lexical}. Since the terms, regardless of their source, can only be utilized as long as they exist in the corpus, the term coverage solely depends on the choice of the corpus. By analyzing the previous setups carefully, we discover that the terms found in WMT are mostly uni- or bi-grams (see Figure \ref{fig:ngram_dist}) and highly colloquial (see Table \ref{tab:dinu_vs_ours} for the top 10 most frequent terms).
These leave the question of whether the previous methods are effective in domain-specific scenarios where accurate terminology translation is truly vital.

In this paper, inspired by the recent masked span prediction models, which have demonstrated improved representation learning capability of contiguous words \cite{song2019mass, joshi2019spanbert, lewis2019bart, raffel2019exploring}, we propose a simple yet effective training scheme to improve terminology translation in highly specialized domains. 
We specifically select two highly specialized domains (\textit{i.e.}, law and medicine) which contain domain-specific terminologies to address more challenging and realistic setups, in addition to applying it to both  typologically similar and dissimilar pairs of languages (German-English (De→En) and Korean-English (Ko→En)).
Thanks to its simplicity, the proposed method is compatible with any autoregressive Transformer-based model, including ones capable of utilizing term dictionaries at training or test time.
In domain-specific setups where longer n-gram terms are pervasive, our method demonstrates improved performance over the standard maximum likelihood estimation (MLE) approach in terms of terminology and sentence-level translation quality. Our code and datasets are available at \url{https://github.com/wns823/NMT_SSP}.

\section{Background}

\paragraph{Lexically constrained NMT} We could group lexically constrained NMT methods into two streams: \textit{hard} and \textit{soft}. The hard approaches aim to force all terminology constraints to appear in the generated output. The methods include replacing constraints \cite{crego2016systran}, constrained decoding \cite{hokamp2017lexically, chatterjee2017guiding, post2018fast, hasler2018neural}, and additional attention heads for external supervision \cite{song2020alignment}.
Although those approaches are reliable and widely used in practice, they typically require a pre-specified term dictionary and an extra candidate selection module if there are multiple matching candidates for a single term (see caption in Table \ref{tab:term_characteristic}).

Several soft methods address this problem without the help of a term dictionary, one of which is training on both constraint pseudo-labeled (with statistical MT) and unlabeled data \cite{song2019code}. More recently, \citet{susanto2020lexically} and \citet{chen2020lexical} proposed methods that do not assume any word alignment or dictionary supervision at training time to handle unseen terms at test time. For their flexibility, we choose them as our baselines. As discussed in Section \ref{sec:introduction}, most previous methods are trained and evaluated on general domain corpora. In this work, we instead tackle highly specialized domain-specific corpora such as law and medicine, where the terms are much longer and often rare.

\paragraph{Domain-specific NMT} Another line of research related to our problem is domain-specific NMT, where difficulties arise from both a limited amount of parallel data and specialized lexicons. Similar to the hard approaches in lexically constrained NMT, several works rely on domain-specific dictionaries \cite{zhang2016bridging, hu2019domain, thompson2019hablex, peng2020dictionary} when generating translations, but they are also prone to the same issues. Other domain-specific NMT methods include unsupervised lexicon adaptation \cite{hu2019domain}, synthetic parallel data generation with monolingual data \cite{sennrich2015improving}, and multi-task learning that combines language modeling and translation objectives \cite{gulcehre2015using, zhang2016exploiting, domhan2017using}.
Our method is a form of multi-task learning by utilizing both the source and target language text for an additional task, while the previous works mostly use only the target language text.

\begin{table}[t!]
\centering
\renewcommand{\arraystretch}{1.5} 
\begin{adjustbox}{width=\columnwidth,center}
\begin{tabular}{c|l} 
\hline
\hline
\multicolumn{2}{c}{\textbf{\citet{dinu2019training}'s dataset}} \\ 
\hline
\hline
\textbf{Iate-414} & gold(15), CDU(13), bridge(12), China(11), Syria(11), night(11), \\
\multicolumn{1}{l|}{} & campaign(11), generation(9), month(7), Iraq(7) \\ 
\hline
\textbf{Iate-581} & gold(26), doping(23), CDU(19), sport(17), US(15), bridge(14), \\
\multicolumn{1}{l|}{} & Syria(13), campaign(13), China(11), night(11) \\ 
\hline
\textbf{Wikt-727} & percent(61), police(50), Thursday(41), Putin(19), five(17), \\
\multicolumn{1}{l|}{} & September(14), Venus(13), later(12), Tuesday(11), less(11) \\ 
\hline
\textbf{Wikt-975} & percent(61), police(59), Thursday(44), Putin(24), old(21), \\
\multicolumn{1}{l|}{} & September(21), five(16), swimmer(14), later(13), Venus(13) \\ 
\hline
\multicolumn{2}{c}{}  \\
\hline
\hline
\multicolumn{2}{c}{\textbf{Our dataset}} \\ 
\hline\hline
\textbf{Law (De-En)} & Council(706), Regulation(521), Commission(481), Union(478), \\
\multicolumn{1}{l|}{} & Treaty(345), Official Journal(319), Member State(283), proposal(239), \\
\multicolumn{1}{l|}{} & on a proposal from the Commission(229), market(181) \\ 
\hline
\textbf{Medical (De-En)} & injection(469), water(275), water for injection(270), patient(269), \\
\multicolumn{1}{l|}{} & infusion(258), solution for infusion(226), sodium(159), \\
\multicolumn{1}{l|}{} & distribution(127), volume of distribution(125), treatment(120) \\ 
\hline
\textbf{Law (Ko-En)} & si(451), official(445), public official(436), member(436), term of \\
\multicolumn{1}{l|}{} & office(367), gu(265), education(209), period(180), term(180), \\
\multicolumn{1}{l|}{} & management(156) \\
\hline
\end{tabular}
\end{adjustbox}
\caption{Top 10 most frequent terms in \citet{dinu2019training}'s and our test splits. Numbers in parenthesis indicate the frequency of terms in each data. As shown in the two tables, all top 10 terms in the WMT corpus are unigrams, while there are longer terms (up to 6-grams) in the domain-specific corpora. Furthermore, compared to WMT, the terms in the domain-specific corpora are more specialized for their corresponding domains.}
\label{tab:dinu_vs_ours}
\end{table}


\paragraph{Span-based Masking} Span-based masking is to predict the spans of masked tokens, as opposed to individual token predictions in BERT \cite{devlin2018bert}.
With this training objective, the model showed improved performance on span-level tasks including question answering and coreference resolution  \cite{joshi2019spanbert}. Concurrently, autoregressive sequence-to-sequence pre-trained models also utilized span-based masking as their objectives and demonstrated its effectiveness in many downstream tasks \cite{song2019mass, lewis2019bart, raffel2019exploring}.
Similar to theirs, our training scheme takes advantage of autoregressive span-based prediction but we condition on both the source language and the previous non-masked target language tokens.

\section{Approach}
\subsection{Source-Conditioned Masked Span Prediction}
We posit that adopting auxiliary span-level supervision in generation can benefit both short and long terminology and sentence-level translation. We, therefore, propose an extra span-level prediction task in translation|namely, source-conditioned masked span prediction (SSP). Different from the recent sequence-to-sequence pre-trained models \cite{song2019mass, lewis2019bart, raffel2019exploring}, our approach applies span masking only on the target side. By conditioning on the full context of the source language and the previous non-masked target language tokens (due to autoregressive decoding), the model is forced to predict the spans of missing tokens given fully referenced information in the encoder and partially in the decoder.

\paragraph{Span masking} We follow the masking procedure proposed in SpanBERT \cite{joshi2019spanbert}, where we first sample the length of spans from a clamped geometric distribution (p=0.2, max=10) and then corrupt 80\% of masked tokens with [MASK], 10\% with random tokens, and 10\% unchanged. We set the corruption ratio to 50\%.


\paragraph{Multi-task Learning} As our training scheme consists of two objectives (\textit{i.e.} translation and masked span prediction), we define the total training objective as follows. Let $\theta$ be the model parameter and $\mathbb{C}$ be the term-matched corpus where each sentence contains at least one or more terms.
The first objective, translation, is to maximize the likelihood of the conditional probability of $y$:

\begin{equation}
p_{\theta}(y|x) = \prod^{T+1}_{t=1}p_{\theta}(y_{t}|y_{0:t-1},x),
\label{eq:trans_loss}
\end{equation}

\noindent where $y=(y_{1},\mathellipsis,y_{T})$ is the target ground-truth (GT) sentence with length $T$ and $x=(x_{1},\mathellipsis,x_{S})$ is the source sentence with length $S$. For the SSP objective, we first corrupt random spans of $y$ until the corruption ratio, resulting in $\tilde{y}$. Then we autoregressively predict the masked tokens $\bar{y}$ while conditioned on both $\tilde{y}$ and $x$:

\begin{equation}
p_{\theta}(\bar{y}|\tilde{y},x) = \prod^{T+1}_{t=1} m_{t} p_{\theta}(y_{t}|\tilde{y}_{0:t-1},x),
\label{eq:ssp_loss}
\end{equation}

\noindent where $m_{t} = 1$ means $y_{t}$ is masked. 

Finally, we simultaneously optimize the joint loss:

\begin{equation}
\begin{aligned}
\mathcal{L}_{total} = - \frac{1}{|\mathbb{C}|} \sum_{(x,y) \sim  \mathbb{C}, \atop \tilde{y} \sim C(y)} & \log{}p_{\theta}(y|x) \\
+\gamma & \log{}p_{\theta}(\bar{y}|\tilde{y},x),
\end{aligned}
\label{eq:total_loss}
\end{equation}

\noindent where $C$ is a span-level corrupter and $\gamma$ is a task coefficient that weights the relative contribution of SSP.

\begin{table}[t!]
\centering
\renewcommand{\arraystretch}{2}
\begin{adjustbox}{width=\columnwidth,center}
\begin{tabular}{c|c||c|c|c|c|c} 
\cline{3-7}
\multicolumn{1}{c}{}& & 
\textbf{\# Sent.} 
& \textbf{\makecell[c]{Avg. words \\ per sent.}}
& \textbf{\# Terms} 
& \textbf{\makecell[c]{Avg. terms \\ per sent.}} 
& \textbf{\makecell[c]{\# Unique \\ terms}} \\ 
\hline
\hline
\multirow{2}{*}{\makecell[c]{Law \\ (De→En) }} & SRC & 
\multirow{2}{*}{\makecell[c]{447,410 \\ (441K/3K/3K)}}
& 27.46 
& \multirow{2}{*}{1,677,852}
& \multirow{2}{*}{2.33}
& 25,460 \\ 
\cline{2-2}\cline{4-4}\cline{7-7}
& TRG & 
& 30.77 
& 
& 
& 27,755 \\ 
\hline
\hline
\multirow{2}{*}{\makecell[c]{Medical \\ (De→En) }} & SRC & 
\multirow{2}{*}{\makecell[c]{494,316 \\ (488K/3K/3K)}}
& 19.01
& \multirow{2}{*}{1,494,269}
& \multirow{2}{*}{1.34}
& 8,633 \\ 
\cline{2-2}\cline{4-4}\cline{7-7}
& TRG & 
& 20.25 
&
& 
& 8,990 \\ 
\hline
\hline
\multirow{2}{*}{\makecell[c]{Law \\ (Ko→En) }} & SRC & 
\multirow{2}{*}{\makecell[c]{93,240 \\ (87K/3K/3K)}} 
& 16.52
& \multirow{2}{*}{353,894}
& \multirow{2}{*}{3.52}
& 2,354 \\ 
\cline{2-2}\cline{4-4}\cline{7-7}
& TRG & 
& 34.56 
& 
& 
& 2,733 \\
\hline
\end{tabular}
\end{adjustbox}
\caption{Statistics of the filtered corpus and matched terms. Note that \textit{\# unique terms} in the source (SRC) and target (TRG) languages are not the same. For instance, ``Arzneimittel'' can translate into multiple forms|``pharmaceutical products'', ``drug'', ``medicinal product'', etc.| depending on the context.}
\label{tab:term_characteristic}
\end{table}

\section{Experiments}

\begin{table*}[!t]
\centering
\renewcommand{\arraystretch}{1.5}
\resizebox{\textwidth}{!}{\begin{tabular}{l||c|cccc|cc|c|cccc|cc}
\hline
\multirow{3}{*}{\textbf{ Model } }    & \multicolumn{7}{c|}{\textbf{Law (De→En)}} & \multicolumn{7}{c}{\textbf{Law (Ko→En)}}      \\ 
\cline{2-15}
        & \multirow{2}{*}{\textbf{BLEU}} & \multicolumn{4}{c|}{\textbf{Term\% $(\uparrow)$}} & \multicolumn{2}{c|}{\textbf{LSM-2 $(\uparrow)$}}        & \multirow{2}{*}{\textbf{BLEU}} & \multicolumn{4}{c|}{\textbf{Term\% $(\uparrow)$}}   & \multicolumn{2}{c}{\textbf{LSM-2 $(\uparrow)$}}                         \\ 
\cline{3-8}\cline{10-15}
 &   & \multicolumn{1}{c}{1-gram} & \multicolumn{1}{c}{2-gram} & \multicolumn{1}{c}{2\textgreater{}micro} & \multicolumn{1}{c|}{2\textgreater{}macro} & \multicolumn{1}{c}{2\textgreater{}micro} & \multicolumn{1}{c|}{2\textgreater{}macro} &   & \multicolumn{1}{c}{1-gram} & \multicolumn{1}{c}{2-gram} & \multicolumn{1}{c}{2\textgreater{}mirco} & \multicolumn{1}{c|}{2\textgreater{}marco} & \multicolumn{1}{c}{2\textgreater{}micro} & \multicolumn{1}{c}{2\textgreater{}macro}  \\ 

\hline
\hline

\textsc{gu19}
 & 70.64 & 
 \multicolumn{1}{c}{94.36}       &
 \multicolumn{1}{c}{92.33}       &
 \multicolumn{1}{c}{73.31}       &
 \multicolumn{1}{c|}{45.25}      &
 \multicolumn{1}{c}{86.22}       &
 \multicolumn{1}{c|}{74.92} 
 & 51.31 &
 \multicolumn{1}{c}{81.47}       &
 \multicolumn{1}{c}{76.51}       &
 \multicolumn{1}{c}{58.15}       &
 \multicolumn{1}{c|}{38.51}      &
 \multicolumn{1}{c}{69.41}       &
 \multicolumn{1}{c}{62.52}        \\ 
 
 \hline 

 \textsc{Vaswani17} 
 & 75.24 & 
 \multicolumn{1}{c}{95.80}       & 
 \multicolumn{1}{c}{93.87}       & 
 \multicolumn{1}{c}{80.29}       & 
 \multicolumn{1}{c|}{55.31}      & 
 \multicolumn{1}{c}{89.71}       & 
 \multicolumn{1}{c|}{79.77}      
 & 53.01 &
 \multicolumn{1}{c}{84.97}       & 
 \multicolumn{1}{c}{81.29}       & 
 \multicolumn{1}{c}{65.79}       & 
 \multicolumn{1}{c|}{54.55}      & 
 \multicolumn{1}{c}{74.29}       & 
 \multicolumn{1}{c}{70.56}        \\ 
 
 \textsc{\quad+SSP}
 & \textbf{75.44} & 
 \multicolumn{1}{c}{\textbf{96.01}}       & 
 \multicolumn{1}{c}{94.08}       & 
 \multicolumn{1}{c}{\textbf{81.52}}       & 
 \multicolumn{1}{c|}{\textbf{58.79}}      & 
 \multicolumn{1}{c}{\textbf{90.81}}       & 
 \multicolumn{1}{c|}{\textbf{82.50}}      
 & \textbf{53.80} &
 \multicolumn{1}{c}{\textbf{85.84}}       & 
 \multicolumn{1}{c}{\textbf{83.94}}       & 
 \multicolumn{1}{c}{\textbf{66.84}}       & 
 \multicolumn{1}{c|}{\textbf{58.15}}      & 
 \multicolumn{1}{c}{75.71}       & 
 \multicolumn{1}{c}{69.82}        \\ 
\hline

 \textsc{Chen20} 
 & 74.19 & 
 \multicolumn{1}{c}{95.55}       & 
 \multicolumn{1}{c}{94.08}       & 
 \multicolumn{1}{c}{80.63}       & 
 \multicolumn{1}{c|}{54.73}      & 
 \multicolumn{1}{c}{89.80}       & 
 \multicolumn{1}{c|}{80.89}      
 & 53.08 &
 \multicolumn{1}{c}{85.49}       & 
 \multicolumn{1}{c}{83.25}       & 
 \multicolumn{1}{c}{65.51}       & 
 \multicolumn{1}{c|}{52.49}      & 
 \multicolumn{1}{c}{74.53}       & 
 \multicolumn{1}{c}{67.81}        \\ 

 \textsc{\quad+SSP} 
 & 75.24 & 
 \multicolumn{1}{c}{95.92}       & 
 \multicolumn{1}{c}{\textbf{94.50}}       & 
 \multicolumn{1}{c}{81.31}       & 
 \multicolumn{1}{c|}{56.33}      & 
 \multicolumn{1}{c}{90.40}       & 
 \multicolumn{1}{c|}{81.91}      
 & 53.32 &
 \multicolumn{1}{c}{85.63}       & 
 \multicolumn{1}{c}{82.10}       & 
 \multicolumn{1}{c}{66.19}       & 
 \multicolumn{1}{c|}{56.50}      & 
 \multicolumn{1}{c}{\textbf{76.02}}       & 
 \multicolumn{1}{c}{\textbf{72.27}}        \\ 
\hline

\end{tabular}}
\caption{(\textit{Without dictionary}) Results on legal domain corpora (De→En and Ko→En) without terminology guidance at test time. \textsc{Vaswani17} combined with our training objective (Eq.\eqref{eq:total_loss}) outperforms other methods in most cases. Note that \textsc{gu19} is a non-autoregressive model, therefore not applicable to our proposed method. Higher Term\% and LSM-2 mean better performance.}
\label{tab:law_result}
\end{table*}

\begin{table*}[!t]
\centering
\renewcommand{\arraystretch}{1.5}
\resizebox{\textwidth}{!}{\begin{tabular}{l||c|cccc|cc|c|cccc|cc}
\hline
\multirow{3}{*}{\textbf{ Model } }    & \multicolumn{7}{c|}{\textbf{Law (De→En)}} & \multicolumn{7}{c}{\textbf{Law (Ko→En)}}      \\ 
\cline{2-15}
        & \multirow{2}{*}{\textbf{BLEU}} & \multicolumn{4}{c|}{\textbf{Term\% $(\uparrow)$}} & \multicolumn{2}{c|}{\textbf{LSM-2 $(\uparrow)$}}        & \multirow{2}{*}{\textbf{BLEU}} & \multicolumn{4}{c|}{\textbf{Term\% $(\uparrow)$}}   & \multicolumn{2}{c}{\textbf{LSM-2 $(\uparrow)$}}                         \\ 
\cline{3-8}\cline{10-15}
 &   & \multicolumn{1}{c}{1-gram} & \multicolumn{1}{c}{2-gram} & \multicolumn{1}{c}{2\textgreater{}micro} & \multicolumn{1}{c|}{2\textgreater{}macro} & \multicolumn{1}{c}{2\textgreater{}micro} & \multicolumn{1}{c|}{2\textgreater{}macro} &   & \multicolumn{1}{c}{1-gram} & \multicolumn{1}{c}{2-gram} & \multicolumn{1}{c}{2\textgreater{}mirco} & \multicolumn{1}{c|}{2\textgreater{}marco} & \multicolumn{1}{c}{2\textgreater{}micro} & \multicolumn{1}{c}{2\textgreater{}macro}  \\ 

\hline
\hline

\textsc{Susanto20} 
 & 62.20 & 
 \multicolumn{1}{c}{94.38}       & 
 \multicolumn{1}{c}{92.95}       &
 \multicolumn{1}{c}{82.06}       &
 \multicolumn{1}{c|}{\textbf{64.06}}      &
 \multicolumn{1}{c}{\textbf{94.93}}       &
 \multicolumn{1}{c|}{\textbf{92.14}} 
 & 50.56 &
 \multicolumn{1}{c}{81.67}       &
 \multicolumn{1}{c}{76.74}       &
 \multicolumn{1}{c}{58.47}       &
 \multicolumn{1}{c|}{38.66}      &
 \multicolumn{1}{c}{69.63}       &
 \multicolumn{1}{c}{62.60}        \\ 
\hline
 \textsc{Chen20} 
 & 73.05 & 
 \multicolumn{1}{c}{96.64}       & 
 \multicolumn{1}{c}{93.29}       & 
 \multicolumn{1}{c}{78.73}       & 
 \multicolumn{1}{c|}{51.47}      & 
 \multicolumn{1}{c}{90.00}       & 
 \multicolumn{1}{c|}{80.29}      
 & 52.60 &
 \multicolumn{1}{c}{84.74}       & 
 \multicolumn{1}{c}{83.94}       & 
 \multicolumn{1}{c}{67.33}       & 
 \multicolumn{1}{c|}{59.53}      & 
 \multicolumn{1}{c}{75.59}       & 
 \multicolumn{1}{c}{74.54}        \\ 

 \textsc{\quad+SSP} 
 & \textbf{74.72} & 
 \multicolumn{1}{c}{\textbf{97.15}}       & 
 \multicolumn{1}{c}{\textbf{95.95}}       & 
 \multicolumn{1}{c}{\textbf{84.67}}       & 
 \multicolumn{1}{c|}{57.48}      & 
 \multicolumn{1}{c}{93.94}       & 
 \multicolumn{1}{c|}{83.62}      
 & \textbf{53.38} &
 \multicolumn{1}{c}{\textbf{95.86}}       & 
 \multicolumn{1}{c}{\textbf{94.92}}       & 
 \multicolumn{1}{c}{\textbf{88.58}}       & 
 \multicolumn{1}{c|}{\textbf{79.34}}      & 
 \multicolumn{1}{c}{\textbf{94.17}}       & 
 \multicolumn{1}{c}{\textbf{91.48}}        \\ 
\hline

\end{tabular}}
\caption{(\textit{With dictionary}) Results on legal domain corpora when the GT terms are provided at test time. +SSP consistently shows improvements over its MLE counterparts. Contrary to the previous findings \cite{susanto2020lexically, chen2020lexical}, the models do not show improved BLEU scores compared to those in Table \ref{tab:law_result}. We argue that providing terms at test time is indeed helpful for terminology generation, but it can often hinder the generation of fluent text. This becomes more apparent in our non-autoregressive setup.}
\label{tab:law_dict_result}
\end{table*}

\subsection{Setup}
\label{sec:experiment}

\noindent \textbf{Data} \enskip We use De-En legal and medical domain corpora from OPUS\footnote{\url{http://opus.nlpl.eu/}} \cite{tiedemann2012parallel} and the De-En bilingual term dictionary from IATE\footnote{\url{https://iate.europa.eu}}. Terms in different languages are aligned via term IDs. For the typologically distant pair of languages, we use the Ko-En legal domain corpus available on AI Hub\footnote{\url{https://www.aihub.or.kr/aidata/87}}, and the manually processed bilingual term dictionary downloaded from the Korea Legislation Research Institute (KLRI) website\footnote{\url{https://www.moleg.go.kr/board.es?mid=a10504000000&bid=0010&act=view&list_no=43927&nPage=2}}. In cases where one term translates into multiple terms, we consider all possible pairs to maximize the number of sentence and term matches. 

Among the selected pairs, we further filtered out terms that are less than four characters long and longer than 20 grams. Sentences that do not contain any term are also removed. 
The statistics of the datasets are reported in Table \ref{tab:term_characteristic}.
More details about the preprocessing steps are in Appendix \ref{appendix:preprocessing}.

For data splitting, we developed a new data splitting algorithm that considers the same distribution of n-grams across each data split. We use 3,000 sentences for valid and test sets in case of high redundancy in certain corpora, while previous works that utilize OPUS use only 2,000 \cite{koehn-knowles-2017-six, muller2019domain}. It is important to note that all the sentences in our data splits are matched with domain-specific terms (i.e. at least one or more terms exist in each sentence) following the style of \citet{dinu2019training}. The pseudo-code for the terminology-aware data split algorithm is in Appendix \ref{appendix:data_split}.

\paragraph{Baselines} We compare our method on two recent lexically constrained NMT models of different natures: autoregressive \cite{chen2020lexical} and non-autoregressive \cite{susanto2020lexically}, but both can operate with or without a term dictionary at test time. We refer to them as \textsc{Chen20} and \textsc{Susanto20}, respectively. +SSP indicates models trained with our proposed training scheme, while no indication is the standard MLE method. A base Transformer \cite{vaswani2017attention}, denoted as \textsc{Vaswani17}, and a Levenshtein Transformer \cite{gu2019levenshtein}, denoted as \textsc{gu19}, are also reported to compare the relative performance between models. \textsc{Susanto20} without a dictionary is equivalent to \textsc{gu19}.

\paragraph{Evaluation} We use SacreBLEU\footnote{\url{https://github.com/mjpost/sacrebleu}} \cite{post2018call} for measuring translation quality. For terminology translation, we use \textit{term usage rate} for both short ($\leq$2-grams) and long ($>$2-grams) terms. Term usage rate (Term\%) is the number of generated terms divided by the total number of terms \cite{dinu2019training, susanto2020lexically}. Specifically for evaluating long terms, we report both the macro and micro averages due to the heavy-tailed nature of n-grams.
In addition, although exact term translation is the primary objective for terminology translation, due to its harshness, evaluating models only with Term\% may not fully describe the models' behavior.
Therefore, we also evaluate each model in terms of partial n-gram matches, which is explained in the next paragraph. All evaluations are conducted with a beam size of 5.

\paragraph{Partial N-gram Match} Inspired by the longest common substring problem \cite{gusfield_1997}, we devised a partial n-gram match score for evaluating long terminology|\textit{longest sub n-gram match} (LSM) score. Formally, let the generated target sentence be $\hat{y}=(\hat{y}_{1}, ..., \hat{y}_{T})$ and the matched terms for the target ground truth (GT) sentence $y$ be $y'=\bigcup^{N}_{i=1}(y'_{i1},...,y'_{il})$, where $N$ is the number of GT terms in $y$ and $l$ is an arbitrary n-gram length for $i$-th term. Then, LSM is defined as the ratio of the longest n-gram overlap divided by $l$. As too many overlaps can happen at the uni and bi-gram levels, we only calculate LSM for long terminology ($>$2 grams) and set the least overlap to be greater than or equal to 2 grams, all else being zero, therefore denoted as LSM-2.

\subsection{Results and Analysis}
\label{subsec:result}
For the legal domain, where many terms are exceptionally long compared to most other domains, our training scheme shows consistent improvements over the standard MLE counterparts, as shown in Table \ref{tab:law_result} and Table \ref{tab:law_dict_result}. Even with the extreme setting of law Ko→En, low-resourced and typologically divergent, our method is still effective in most metrics we use.
Compared to the autoregressive models, \textsc{gu19} and \textsc{Susanto20} did not achieve competitive BLEU scores in our domain-specific setup.
We suspect that this is due to both its complex decoding nature and the small amount of training data (originally WMT).
Sampled translation results are reported in Table \ref{tab:qual_ex}.

\begin{table}[t!]
\renewcommand{\arraystretch}{1.5}
\begin{adjustbox}{width=\columnwidth,center}
\begin{tabular}{l||c|cccc|cc} 
\hline
\multirow{3}{*}{\textbf{ Model }} & \multicolumn{7}{c}{\textbf{Medical (De→En)}} \\
\cline{2-8} & \multirow{2}{*}{\textbf{BLEU }} & \multicolumn{4}{c|}{\textbf{Term\% $(\uparrow)$}} & \multicolumn{2}{c}{\textbf{LSM-2 $(\uparrow)$}} \\ 
\cline{3-8} & & \textbf{1-gram} & \textbf{2-gram} & 
\textbf{2\textgreater{}micro} & \textbf{2\textgreater{}macro} & \textbf{2\textgreater{}micro} & \textbf{2\textgreater{}macro}  \\ 

\hline
\hline
\multicolumn{8}{c}{\textit{Without dictionary}} \\
\hline
\textsc{gu19} & 70.85 & 93.83 & 91.24 & 77.46 & 53.66 & 86.15 & 75.21 \\
\hline
\textsc{Vaswani17} & 76.31 & 94.22 & 90.80 & 79.82 & 61.03 & 87.11 & 80.48 \\
\textsc{\quad+SSP} & \textbf{76.87} & 94.36 & \textbf{91.31} & \textbf{80.63} & 53.68 & 88.01 & 74.76 \\
\hline
\textsc{Chen20} & 74.84 & 94.29 & 90.61 & 79.42 & \textbf{68.37} & 87.13 & \textbf{84.64} \\
\textsc{\quad+SSP} & 76.72 & \textbf{94.61} & 90.42 & 80.41 & 68.03 & \textbf{88.08} & 83.04 \\
\hline
\hline

\multicolumn{8}{c}{\textit{With dictionary}} \\

\hline
\textsc{Susanto20} & 62.20 & 91.01 & 92.64 & 88.09 & 67.22 & \textbf{95.46} & \textbf{94.27} \\
\hline
\textsc{Chen20} & 72.84 & 94.40 & 93.58 & 83.77 & 67.98 & 89.95 & 86.70 \\
\textsc{\quad+SSP} & \textbf{75.50} & \textbf{95.86} & \textbf{94.92} & \textbf{88.58} & \textbf{79.34} & 94.17 & 91.48 \\
\hline

\end{tabular}
\end{adjustbox}
\caption{Results on the medical domain dataset (De→En).}
\label{tab:med_result}
\end{table}

For the medical domain, the behaviors of two baselines, \textsc{Vaswani17} and \textsc{Chen20}, are not clearly shown compared to the legal domain. However, our training scheme shows consistent improvements in BLEU and Term\% at 2$>$micro which reflects the global performance of long terminology generation. Similar to the legal De→En results, \textsc{Susanto20} shows better performance on several metrics on long terminology translation, but the BLEU score is decreased by about 8 points, compared to no dictionary use.

\section{Conclusion}
We propose a simple and effective training scheme for improving lexically constrained NMT by introducing the masked span prediction task on the decoder side. Our method shows its effectiveness in terms of terminology and sentence-level translation over the standard MLE training in highly specialized domains in two language pairs. As we publicly release our code and datasets, we hope that more people can join this area of research without much burden. In the future, we plan to further investigate applying our method to non-autoregressive methods.

\section*{Acknowledgments}
We want to thank Minjoon Seo for his constructive comments on our research direction and Wooju Kim for his help in starting this topic. We also thank anonymous reviewers for their effort and valuable feedback. This work was supported by Institute of Information \& Communications Technology Planning \& Evaluation (IITP) grant (No.2019-0-00075, Artificial Intelligence Graduate School Program (KAIST)), and National Research Foundation of Korea (NRF) grant (NRF-2020H1D3A2A03100945), funded by the Korea government (MSIT).


\bibliographystyle{acl_natbib}
\bibliography{anthology,acl2021}

\clearpage

\appendix 
\newpage
\section{Preprocessing and Training}

\subsection{Preprocessing} 
\label{appendix:preprocessing}

For De→En, we applied Moses tokenization \cite{koehn-etal-2007} and joint source-target byte pair encoding (BPE) \cite{sennrich-etal-2016-neural} with 20,000 split operations. For En→Ko, English was tokenized using spaCy\footnote{\url{https://spacy.io/}} and Korean using KoNLPy's MeCab-ko\footnote{\url{https://konlpy.org/en/latest/}} \cite{park2014konlpy}, followed by BPE with 20,000 operations. We apply sentence filtering up to 80 tokens.

\subsection{Training} 

\paragraph{Model} We follow the base Transformer architecture and fix the same hyperparameter configurations for all baselines. For the exact implementation of each baseline, we followed the authors' official code on github (\textsc{Chen20}\footnote{\url{https://github.com/ghchen18/leca}} and \textsc{Susanto20}\footnote{\url{https://github.com/raymondhs/constrained-levt}}). We implemented our code using \textsc{fairseq}\footnote{\url{https://github.com/pytorch/fairseq}} \cite{ott2019fairseq}, trained on Nvidia GeForce RTX 3090 and RTX 2080 Ti GPUs.


\paragraph{Hyperparameter} Detailed hyperparameter settings of baselines are reported below. Span masking and task coefficient only apply to our proposed training scheme.

\begin{table}[h!]
\centering
\begin{small}
\begin{tabular}{c|c}
\hline
\multicolumn{2}{c}{Transformer} \\
\hline
Embedding dim. & 512 \\
Transformer FFN dim. & 2048 \\
Enc/Decoder layers & 6 \\
Attention heads & 8 \\
Share all embedding & True \\
Dropout & 0.3 \\
Label smoothing & 0.1 \\
Optimizer & Adam \\
Learning rate & 0.0005 \\
Warmup updates & 4000 \\
Maximum token per batch & 4096 \\
Maximum token lengths & 80 \\
\hline
\multicolumn{2}{c}{Span Masking} \\
\hline
Span length & Geometric (p=0.2) \\
Maximum span length & 10 \\
Minimum span length & 1 \\
Corruption ratio & 0.5 \\
\hline
\multicolumn{2}{c}{Task Coefficient} \\
\hline
Task coefficient ($\gamma$) & 0.5 \\
\hline
\end{tabular}
\end{small}
\caption{Hyperparameter settings}
\label{tab:hyperparam}
\end{table}

\newpage
\section{Terminology-Aware Data Split Algorithm}

\label{appendix:data_split}
\SetAlFnt{\small}
\begin{algorithm2e}[h!]
\caption{Terminology-Aware Data Split Algorithm}
\KwData{Dictionary $\mathbb{D}$, Corpus $\mathbb{C}$, Held-out Size \textit{R}}
\KwResult{Sent=(Sent$_{train}$, Sent$_{valid}$, Sent$_{test}$) \\
\quad\quad\quad\enspace Term=(Term$_{train}$, Term$_{valid}$, Term$_{test}$)}
\begin{algorithmic}[1]
\STATE Sort {$\mathbb{D}$} in a descending order
\STATE \textit{N} = dict()
\STATE \textit{T'} = dict()
\STATE \textit{S} = ( $|\mathbb{C}|-2$ * \textit{R}, \textit{R}, \textit{R} )
\STATE Sent$_{train}$ = [], Sent$_{valid}$ = [], Sent$_{test}$ = []
\STATE Term$_{train}$ = [], Term$_{valid}$ = [], Term$_{test}$ = []
  \FOR{\textit{i} in \{1,2, ..., $|\mathbb{C}|$\}}
    \STATE x, y = $\mathbb{C}$[\textit{i}]
    \STATE \textit{ngramlist} = []
    \STATE \textit{T''} = []
    \FOR{ (x', y') in {$\mathbb{D}$}}
      \IF { y' in y and x' in x}
        \STATE  \textit{ngramlist}.append(ngram(y'))
        \STATE  y = y.replace(y', ``", 1)
        \STATE  x = x.replace(x', ``", 1)
        \STATE  \textit{T''}.append((x', y')) 

      \ENDIF
    \ENDFOR
    
    \STATE {\textit{n} = Max({\textit{ngramlist}}) }
    \IF {\textit{n} is not in {\textit{N}}.keys() }
    \STATE { {\textit{N}}[{\textit{n}}] = [] }
    \ENDIF
    
    \STATE {\textit{N}}[{\textit{n}}].append(\textit{i})
    \STATE {\textit{T'}}[\textit{i}] = {\textit{T''}}
  \ENDFOR

\STATE {\textit{K}} = Sort the keys in \textit{N} in a descending order.
\FOR{{\textit{k}} in {\textit{K}}}
    \STATE {$i_{dk}$ , $i_{uk}$ = DuplicateCheck($N[k]$)}
    \STATE (Sent, Term) += Distributor\_Dup($i_{dk}$, $N[k]$, $T'$, $S$)
    \STATE (Sent, Term) += Distributor\_Uni($i_{uk}$, $N[k]$, $T'$, $S$)
\ENDFOR

\end{algorithmic}
\label{algo:data_split}
\end{algorithm2e}

\begin{small}
\noindent 
Line 1 : Sort $\mathbb{D}$ w.r.t. target language terms. \\ 
Line 2 : Initialize a dictionary for storing paired sentences. The keys are the longest n-gram lengths for each sentence w.r.t. the target language. \\
Line 3 : Initialize a dictionary for storing matched terms. The keys are the indices of a corresponding sentence. \\
Line 13 : ngram() returns the token length of a term. In our case, it is used for calculating the length of a target language token y'. \\
Line 14 : Replace y' with ``" in y to avoid unwanted substring duplication (e.g., In case of having ``public officer" and ``officer" in a sentence, we would like to first match ``public officer" instead of ``office" when we have ``public officer" in the dictionary. See Line 1). \\
Line 19 : Calculate the maximum length of n-grams in y. \\
Line 23 : Store the sentence index w.r.t. its longest length of n-grams. \\
Line 24 : Store the list of terms w.r.t. its sentence index. \\
Line 28 : DuplicateCheck() checks for duplication in the corpus and returns duplicate and non-duplicate indices. Note that $i_{dk}$ is a list of duplicate sentence indices, and $S_{uk}$ is a list of unique sentence indices. \\
Line 29 : Distributor\_Dup() first calculates the number of sentences and phrases to be distributed across train, valid, and test sets following the ratio in \textit{S}, and then distributes sentences accordingly. \\
Line 30 : Distributor\_Uni() distributes unique sentences and phrases alternatively between train, valid, and test sets. \\
\end{small}

\section{Removing duplicates} As the OPUS datasets contain duplicate sentences \cite{aharoni2020unsupervised}, we further evaluate each model with unseen, unique test samples only.
Similar to Tables \ref{tab:unique_law_result} and \ref{tab:unique_med_result}, our training scheme outperforms its MLE counterparts. The Ko-En law corpus does not contain any duplicate sentence, and therefore the results are equivalent to those in Tables \ref{tab:law_result} and \ref{tab:law_dict_result}.

\begin{table}[h!]
\renewcommand{\arraystretch}{1.5}
\begin{adjustbox}{width=\columnwidth,center}
\begin{tabular}{l||c|cccc|cc} 
\hline
\multirow{3}{*}{\textbf{ Model }} & \multicolumn{7}{c}{\textbf{Law (De→En)}} \\
\cline{2-8} & \multirow{2}{*}{\textbf{BLEU }} & \multicolumn{4}{c|}{\textbf{Term\% $(\uparrow)$}} & \multicolumn{2}{c}{\textbf{LSM-2 $(\uparrow)$}} \\ 
\cline{3-8} & & \textbf{1-gram} & \textbf{2-gram} & 
\textbf{2\textgreater{}micro} & \textbf{2\textgreater{}macro} & \textbf{2\textgreater{}micro} & \textbf{2\textgreater{}macro}  \\ 

\hline
\hline
\multicolumn{8}{c}{\textit{Without dictionary}} \\
\hline
\textsc{gu19} & 68.14 & 93.71 & 91.87 & 72.32 & 46.24 & 85.05 & 74.22 \\
\hline
\textsc{Vaswani17} & 72.86 & 95.30 & 93.36 & 78.68 & 54.99 & 88.40 & 78.69 \\
\textsc{\quad+SSP} & \textbf{73.15} & \textbf{95.57} & 93.54 & \textbf{79.98} & \textbf{59.18} & \textbf{89.64} & \textbf{81.53} \\
\hline
\textsc{Chen20} & 71.89 & 95.04 & 93.54 & 78.83 & 54.55 & 88.43 & 79.63 \\
\textsc{\quad+SSP} & 72.93 & 95.46 & \textbf{94.14} & 79.74 & 56.63 & 89.19 & 80.88 \\
\hline
\hline

\multicolumn{8}{c}{\textit{With dictionary}} \\
\hline
\textsc{Susanto20} & 59.23 & 94.10 & 92.89 & 82.86 & \textbf{68.22} & \textbf{95.28} & \textbf{93.36} \\
\hline
\textsc{Chen20} & 70.90 & 96.36 & 94.01 & 77.26 & 51.97 & 89.12 & 80.06 \\
\textsc{\quad+SSP} & \textbf{72.70} & \textbf{96.85} & \textbf{95.68} & \textbf{83.65} & 58.33 & 93.30 & 83.40 \\
\hline

\end{tabular}
\end{adjustbox}
\caption{Results on the law domain dataset with no duplication in data (De→En).}
\label{tab:unique_law_result}
\end{table}


\begin{table}[h!]
\renewcommand{\arraystretch}{1.5}
\begin{adjustbox}{width=\columnwidth,center}
\begin{tabular}{l||c|cccc|cc} 
\hline
\multirow{3}{*}{\textbf{ Model }} & \multicolumn{7}{c}{\textbf{Medical (De→En)}} \\
\cline{2-8} & \multirow{2}{*}{\textbf{BLEU }} & \multicolumn{4}{c|}{\textbf{Term\% $(\uparrow)$}} & \multicolumn{2}{c}{\textbf{LSM-2 $(\uparrow)$}} \\ 
\cline{3-8} & & \textbf{1-gram} & \textbf{2-gram} & 
\textbf{2\textgreater{}micro} & \textbf{2\textgreater{}macro} & \textbf{2\textgreater{}micro} & \textbf{2\textgreater{}macro}  \\ 

\hline
\hline
\multicolumn{8}{c}{\textit{Without dictionary}} \\
\hline
\textsc{gu19} & 54.27 & 89.93 & 84.35 & 67.83 & 50.44 & 78.18 & 70.33 \\
\hline
\textsc{Vaswani17} & 58.29 & 90.09 & 84.51 & 70.98 & 57.82 & 79.18 & 76.45 \\
\textsc{\quad+SSP} & 59.19 & 90.43 & \textbf{85.50} & \textbf{71.51} & 49.20 & \textbf{80.38} & 70.10 \\
\hline
\textsc{Chen20} & 58.27 & 90.30 & 84.02 & 70.38 & 64.02 & 79.28 & \textbf{80.14} \\
\textsc{\quad+SSP} & \textbf{59.49} & \textbf{90.57} & 83.53 & 71.25 & \textbf{64.36} & 80.29 & 78.70 \\
\hline
\hline

\multicolumn{8}{c}{\textit{With dictionary}} \\
\hline
\textsc{Susanto20} & 45.60 & 88.77 & 89.95 & \textbf{86.86} & 68.22 & \textbf{94.83} & \textbf{93.86} \\
\hline
\textsc{Chen20} & 58.30 & 90.81 & 89.79 & 79.83 & 67.62 & 86.59 & 85.05 \\
\textsc{\quad+SSP} & \textbf{60.30} & \textbf{93.10} & \textbf{91.10} & 85.19 & \textbf{79.26} & 91.34 & 90.51 \\
\hline

\end{tabular}
\end{adjustbox}
\caption{Results on the medical domain dataset with no duplication in data (De→En).}
\label{tab:unique_med_result}
\end{table}

\section{Examples}
Table \ref{tab:qual_ex} shows translation results of the baselines and our method.
\clearpage

\begin{table*}[h]
\renewcommand{\arraystretch}{1.5}
\centering
\resizebox{\textwidth}{!}{
\begin{tabular}{l|l}
\hline
\hline
Source &   dieses Vorbringen wurde zurückgewiesen , da die einschlägigen Bestimmungen der Grundverordnung sehr wohl mit dem WTO-\textbf{Übereinkommen zur Durchführung }  \\
\multicolumn{1}{l|}{}   &   \textbf{des Artikels VI des Allgemeinen Zoll- und Handelsabkommens 1994} und dem \textbf{Übereinkommen über Subventionen und Ausgleichsmaßnahmen} vereinbar sind .  \\ 
\hline

\textsc{Vaswani17} & this claim was rejected because the relevant provisions of the basic Regulation are very compatible with the 1994 WTO Agreement on \\
\multicolumn{1}{l|}{} & Implementation of Article VI of the General Agreement on Tariffs and Trade and the 1994 \textbf{Agreement on Subsidies and Countervailing Measures} . \\
\hdashline
\quad +SSP      & this claim was rejected as the relevant provisions of the basic Regulation are indeed consistent with the WTO \textbf{Agreement on} \\
\multicolumn{1}{l|}{} & \textbf{Implementation of Article VI of the General Agreement on Tariffs and Trade 1994} and the \textbf{Agreement on Subsidies and Countervailing Measures} . \\
\hline
\textsc{Chen20}      & this claim was rejected as the relevant provisions of the basic Regulation are , however , in any event , compatible with the WTO Agreement on \\
\multicolumn{1}{l|}{} & the implementation of Article VI of the General Agreement on Tariffs and Trade 1994 and with the \textbf{Agreement on Subsidies and Countervailing Measures} . \\
\hdashline
\quad +SSP  & this claim was rejected because it is true that the relevant provisions of the basic Regulation are consistent with the \textbf{Agreement on} \\
\multicolumn{1}{l|}{} & \textbf{Implementation of Article VI of the General Agreement on Tariffs and Trade 1994} and the \textbf{Agreement on Subsidies and Countervailing Measures} . \\
\hline
Reference   & this claim was rejected on the grounds that the anti-circumvention provisions of the basic Regulation are not incompatible with the \textbf{Agreement}  \\
\multicolumn{1}{l|}{}   & \textbf{on Implementation of Article VI of the General Agreement on Tariffs and Trade 1994} and the \textbf{ Agreement on Subsidies and Countervailing Measures} .   \\ 

\hline
Terminology &  \makecell[l]{ \{\textbf{Übereinkommen zur Durchführung des Artikels VI des Allgemeinen Zoll- und Handelsabkommens 1994}→\textbf{Agreement on Implementation of Article VI of the General} \\ \textbf{Agreement on Tariffs and Trade 1994}, \textbf{Übereinkommen über Subventionen und Ausgleichsmaßnahmen}→\textbf{Agreement on Subsidies and Countervailing Measure}s\}}          \\ 
\hline
\hline

\multicolumn{2}{l}{} \\

\hline
\hline
Source      &   \makecell[l]{  ( 3 ) Das \textbf{Angebot zur vorzugsweisen Zeichnung} sowie die Frist , innerhalb deren dieses Recht ausgeuebt werden muß, sind Gegenstand einer Bekanntmachung \\ in dem gemäß der Richtlinie 68 / 151 / EWG bestimmten einzelstaatlichen \textbf{Amtsblatt . }} \\
\hline
\textsc{Vaswani17}  &  \makecell[l]{ 3 . the tender for subscription and the time limit within which that right must be exercised shall be published in the \textbf{national gazette} determined in accordance \\ with Directive 68 / 151 / EEC .
} \\
\hdashline
\quad +SSP      &   \makecell[l]{ 3 . the tender for a preference call and the time limit within which that right must be exercised shall be the subject of a notice in the \textbf{national gazette} designated \\ in accordance with Directive 68 / 151 / EEC .
}                             \\
\hline
\textsc{Chen20}      &  \makecell[l]{ 3 . the tender for preferred subscription and the time limit within which it must be exercised shall be the subject of a notice published in the national publication \\ designated pursuant to Directive 68 / 151 / EEC .
}                         \\
\hdashline
\quad +SSP      &   \makecell[l]{ 
3 . tenders for preference drawing together with the time limit within which that right must be exercised shall be the subject of a notice in the \textbf{national gazette} \\ designated in accordance with Directive 68 / 151 / EEC .
}                         \\
\hline
Reference   &   \makecell[l]{ any \textbf{offer of subscription on a pre-emptive basis} and the period within which this right must be exercised shall be published in the \textbf{national gazette} appointed \\ in accordance with Directive 68 / 151 / EEC .}                \\
\hline
Terminology &   \makecell[l]{  \{\textbf{Angebot zur vorzugsweisen Zeichnung}→\textbf{offer of subscription on a pre-emptive basis}, \textbf{Amtsblatt}→\textbf{national gazette}\}  }                         \\ 
\hline
\hline

\multicolumn{2}{l}{} \\

\hline
\hline
Source      &      \makecell[l]{( 19 ) Nach der \textbf{Rechtsprechung} des \textbf{Gerichtshofs} sind einzelstaatliche Vorschriften betreffend die Fristen für die Rechtsverfolgung \textbf{zulässig} , sofern sie für \\ derartige Klagen nicht ungünstiger sind als für gleichartige Klagen , die das innerstaatliche Recht betreffen , und sofern sie die Ausübung der durch \\ das \textbf{Gemeinschaftsrecht} gewährten Rechte nicht praktisch unmöglich machen .}                          \\
\hline
\textsc{Vaswani17} & \makecell[l]{
( 19 ) According to the case law of the \textbf{Court of Justice} , national rules concerning time limits for bringing actions may be allowed , \\ provided that such actions are not less favourable than those relating to the like actions under national law and \\ do not make it impossible in practice to exercise the rights conferred by \textbf{Community law} .
}  \\
\hdashline
\quad +SSP      &   \makecell[l]{
( 19 ) According to the case law of the \textbf{Court of Justice} , national rules concerning the time limits for prosecution are \textbf{admissible} , \\ provided that they are not less favourable to such actions than those for similar actions under national law if they do not make it impossible \\ to exercise the rights conferred by \textbf{Community law} in practice .
}                             \\
\hline
\textsc{Chen20}      &     \makecell[l]{
( 19 ) According to the case law of the \textbf{Court of Justice} , national rules on the time limits for the exercise of jurisdiction may be allowed , \\ provided that they are not less favourable for such actions than for the same actions covered by national law and \\ do not make it impossible in practice to exercise the rights conferred by \textbf{Community law} .
}                           \\
\hdashline
\quad +SSP      &      \makecell[l]{
( 19 ) The \textbf{Court of Justice} has \textbf{case-law} that national provisions relating to time limits for bringing actions may be accepted , \\ provided that they are no less favourable in such actions than those relating to similar actions brought under national law , \\ provided that they do not practically make it impossible for the exercise of rights conferred by \textbf{Community law} .
}         \\
\hline
Reference   &            \makecell[l]{ ( 19 ) According to the \textbf{case-law} of the \textbf{Court of Justice } , national rules relating to time limits for bringing actions are \textbf{admissible} provided that they are not less \\  favourable than time limits for similar actions of a domestic nature and that they do not render the exercise of rights conferred by the \textbf{Community law} impossible in practice .
 }   \\
\hline
Terminology &        \makecell[l]{ \{\textbf{Gerichtshof}→\textbf{Court of Justice}, \textbf{Gemeinschaftsrecht}→\textbf{Community law}, \textbf{zulässig}→\textbf{admissible}, \textbf{Rechtsprechung}→\textbf{case-law}\} \\ }                        \\ 
\hline
\hline
\end{tabular}
}
\caption{Translation outputs of the models trained with or without our method.}
\label{tab:qual_ex}
\end{table*}

\end{document}